\documentclass[11pt, a4paper]{article}
\usepackage[utf8]{inputenc}
\usepackage[T1]{fontenc} 
\usepackage[english]{babel} 
\usepackage{geometry}          
\usepackage{parskip}
\usepackage{multirow}

\usepackage{amsmath, amssymb, amsthm} 
\usepackage{bm}                       
\usepackage{booktabs}

\usepackage[ruled,vlined]{algorithm2e}
\usepackage{graphicx}              
\usepackage{xcolor}                  
\usepackage[colorlinks=true, allcolors=blue]{hyperref} 

\usepackage{eso-pic}
\AddToShipoutPictureBG{%
  \AtPageUpperLeft{%
    \hspace{\dimexpr\paperwidth-1in-1.2cm\relax}%
    \raisebox{-2cm}{\includegraphics[height=2cm]{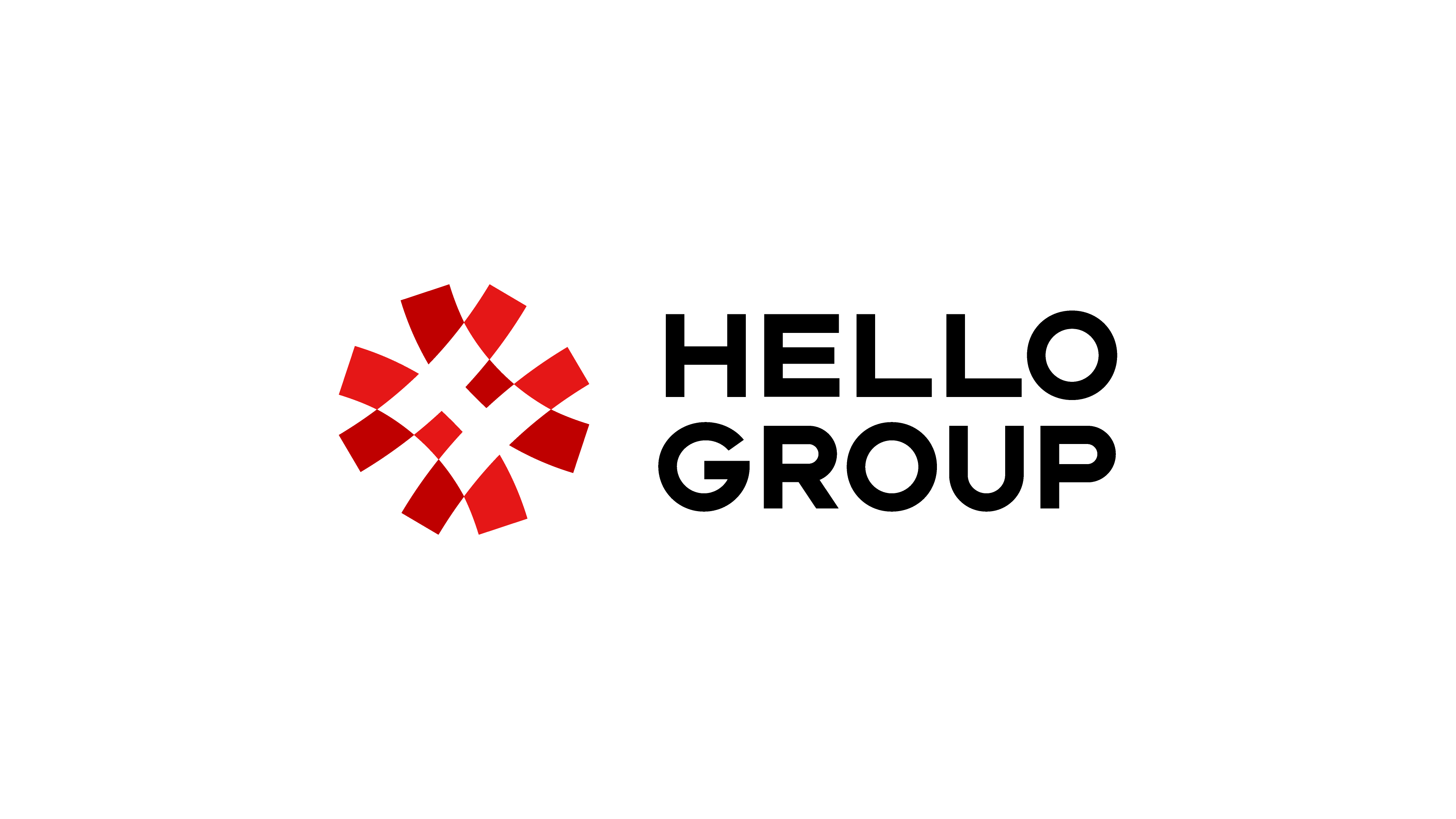}}%
  }%
}

\AddToShipoutPictureBG*{%
  \AtPageLowerLeft{%
    \hspace{1in}%
    \raisebox{0.8 cm}{%
      \parbox{\dimexpr\paperwidth-2in\relax}{%
        \rule{\dimexpr\paperwidth-2in\relax}{0.4pt}\par\nointerlineskip\vskip2pt%
        {\small\url{https://huggingface.co/hellogroup-opensource/AMBER-IMAGE}}%
      }%
    }%
  }%
}
\usepackage[style=authoryear,backend=biber,maxcitenames=2]{biblatex}
\title{\textbf{Amber-Image: Efficient Compression of Large-Scale Diffusion Transformers}}

\author{\texttt{Computational Intelligence Dept, HelloGroup Inc.}}

\date{}

\begin{document}

\maketitle

\begin{abstract}
\noindent
Diffusion Transformer (DiT) architectures have significantly advanced Text-to-Image (T2I) generation but suffer from prohibitive computational costs and deployment barriers. To address these challenges, we propose an efficient compression framework that transforms the 60-layer dual-stream MMDiT-based Qwen-Image into lightweight models without training from scratch. Leveraging this framework, we introduce Amber-Image, a series of streamlined T2I models. We first derive Amber-Image-10B using a timestep-sensitive depth pruning strategy, where retained layers are reinitialized via local weight averaging and optimized through layer-wise distillation and full-parameter fine-tuning. Building on this, we develop Amber-Image-6B by introducing a hybrid-stream architecture that converts deep-layer dual streams into a single stream initialized from the image branch, further refined via progressive distillation and lightweight fine-tuning. Our approach reduces parameters by $70 \%$ and eliminates the need for large-scale data engineering. Notably, the entire compression and training pipeline---from the 10B to the 6B variant---requires fewer than 2{,}000 GPU hours, demonstrating exceptional cost-efficiency compared to training from scratch. Extensive evaluations on benchmarks like DPG-Bench and LongText-Bench show that Amber-Image achieves high-fidelity synthesis and superior text rendering, matching much larger models.

\end{abstract}


\begin{figure}[!htbp]
    \centering
    \includegraphics[width=0.9\textwidth]{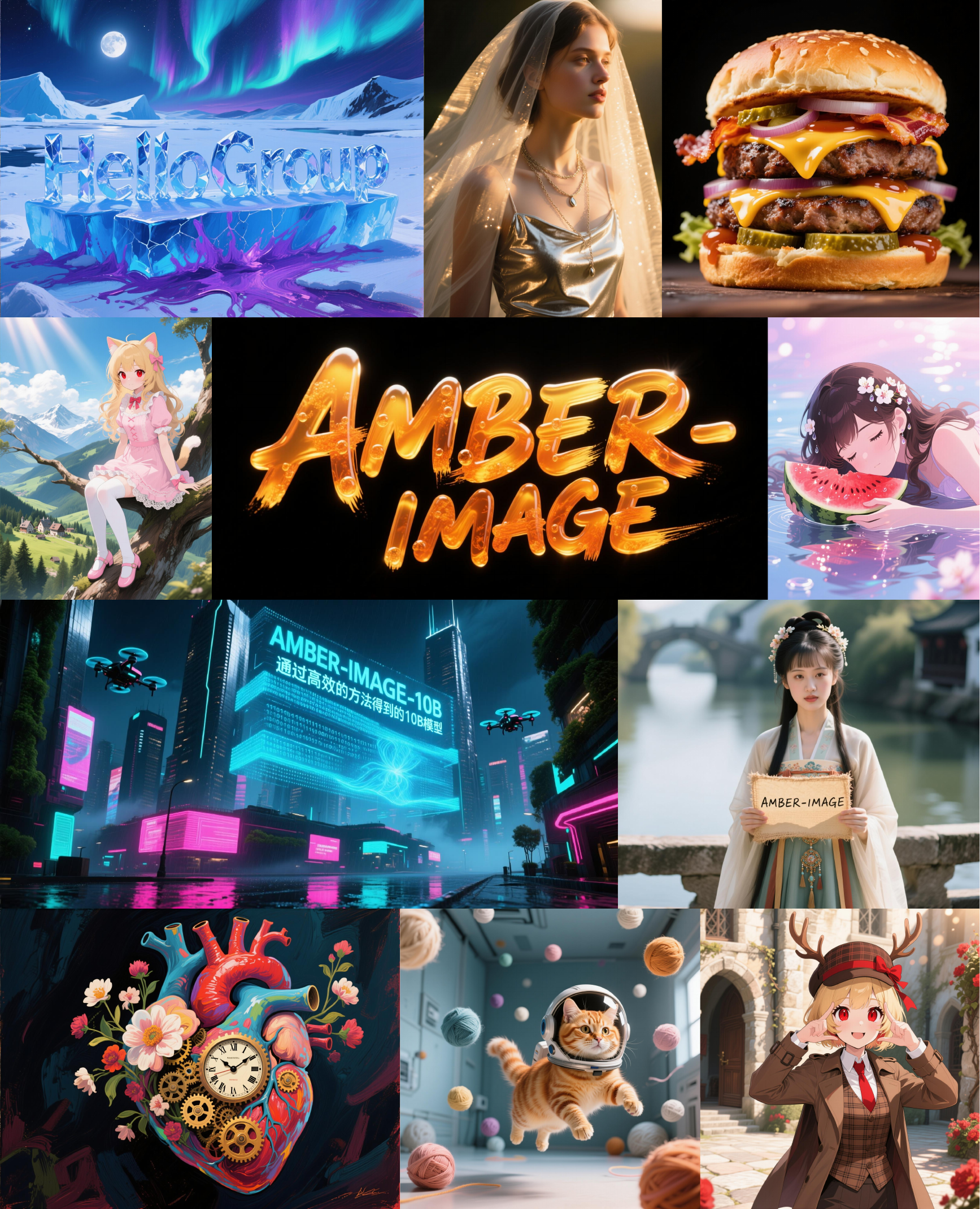}
    \vspace{-0.5em}
    \caption{Representative samples generated by Amber-Image.}
    \vspace{-0.5em}
    \label{fig1}
\end{figure}

\section{Introduction}
In recent years, Text-to-Image (T2I) generation has undergone a revolutionary transformation, evolving from early U-Net-based diffusion models \parencite{rombach2022latent,saharia2022imagen,nichol2022glide} (e.g., Stable Diffusion XL \parencite{podell2023sdxl}) to architectures centered around the Diffusion Transformer (DiT) \parencite{peebles2023sdit,chen2023pixartalpha,chen2024pixartsigma,li2024hunyuandit,xie2024sana}. A pivotal advancement is the Multi-Modal Diffusion Transformer (MMDiT) introduced by Stable Diffusion 3 \parencite{esser2024sd3}, which encodes images and text into a unified sequence of latent tokens. By leveraging deep transformer stacks for joint cross-modal reasoning, MMDiT has significantly elevated the standards for semantic consistency, aesthetic quality, and complex text rendering in synthesized imagery.

However, these advancements have introduced a stark polarization within the current landscape. On the one hand, state-of-the-art closed-source systems, such as Seedream 4 \parencite{seedream2025seedream40} and Nano Banana Pro \parencite{deepmind2025gemini3pro}, deliver exceptional synthesis quality; however, they often lead to significant vendor lock-in. Their proprietary nature restricts fine-grained customization, while exorbitant usage costs pose a significant barrier to practical deployment and widespread adoption. On the other hand, high-performing open-source alternatives (e.g., Qwen-Image \parencite{wu2025qwenimage}, FLUX.2 \parencite{blackforest2025flux2}, and Hunyuan-Image-3.0 \parencite{cao2026hunyuanimage30}) typically rely on massive architectures with tens of billions of parameters, resulting in prohibitive computational demands for both training and inference. Deployment on consumer-grade hardware remains extremely challenging. Although recent efforts have explored training lightweight models from scratch to balance efficiency and performance \parencite{imageteam2025zimage, meituan2025longcat,wang2025ovisimage}, these approaches often require ultra-large-scale, meticulously curated datasets, sophisticated data pipelines, and extensive computational resources—barriers that remain insurmountable for most individual researchers.

In this work, we introduce Amber-Image, a family of efficient text-to-image (T2I) generation models designed to address the prohibitive computational costs and deployment barriers associated with contemporary large-scale generative architectures. Amber-Image achieves high-fidelity synthesis through a dedicated compression pipeline that integrates structured pruning, architectural evolution, and knowledge distillation.

We first derive Amber-Image-10B from Qwen-Image—a 60-layer, 20B-parameter dual-stream MMDiT model where image and text features are processed through separate transformer streams. Using a robust layer importance estimation mechanism, we identify and remove the 30 least critical layers, reducing the parameter count to approximately 10B (Amber-Image-10B). The retained layers are initialized via arithmetic averaging of their pruned neighboring blocks, followed by layer-wise knowledge distillation from the original Qwen-Image model and a brief full-parameter fine-tuning phase on a curated set of high-quality data to recover visual fidelity.

Building upon this foundation, we further propose Amber-Image-6B, which introduces a hybrid-stream architecture: the first 10 layers retain dual-stream processing for modality-specific feature extraction, while the deeper 20 layers are converted to a single stream initialized from the image branch, exploiting cross-modal redundancy. We leverage the trained Amber-Image-10B as a teacher to transfer high-level semantic and aesthetic knowledge via local knowledge distillation, followed by a short, data-efficient fine-tuning stage to align latent distributions and stabilize inference.

Unlike recent lightweight models that rely on training from scratch or massive data engineering, Amber-Image operates entirely through strategic compression and refinement of existing foundation models. Notably, the entire training pipeline---encompassing layer importance estimation, knowledge distillation, and fine-tuning for both the 10B and 6B variants---requires fewer than 2{,}000 GPU hours (using 8 NVIDIA A100 GPUs over approximately 10 days), a fraction of the cost typically associated with training comparable models from scratch. This approach dramatically reduces both computational budget and data requirements while maintaining generation quality competitive with much larger open-source and proprietary systems.

The key contributions of Amber-Image are summarized as follows:

\begin{itemize}
    \item \textbf{Structured Depth Pruning with Fidelity-Aware Initialization:} We propose a layer importance estimation method that accounts for global fidelity impact and timestep sensitivity, enabling safe removal of half the layers in a 60-layer MMDiT. Crucially, we initialize the retained layers via arithmetic averaging of pruned neighboring blocks—a simple yet effective strategy that provides a high-quality warm start and mitigates representational collapse.
    \item \textbf{Progressive Architectural Simplification via Deep-Layer Single-Stream Conversion:} Building on the pruned model, we introduce a hybrid-stream architecture: early layers retain dual-stream processing for modality-specific feature extraction, while deeper layers are converted to a single stream initialized from the image branch. This design further reduces the parameters by $40\%$ with minimal quality loss.
    \item \textbf{Two-Stage Knowledge Transfer without Large-Scale Data:} Our pipeline employs (i) layer-wise distillation from the original full model to recover pruning-induced degradation, followed by (ii) distillation from the intermediate pruned model (Amber-Image-10B) to align the deep single-stream layers. Both stages require only limited fine-tuning on a small, high-quality dataset—eliminating the need for training from scratch or billion-scale data curation.
\end{itemize}

\section{Methodology}

We introduce Amber-Image, an efficient compression framework for large-scale Diffusion Transformers. Built upon Qwen-Image, Amber-Image reuses its text encoder (Qwen2.5-VL-7B) and VAE with official weights unchanged. Our compression targets exclusively the 60-layer dual-stream MMDiT backbone, where image and text latents are processed through separate transformer streams.
\begin{figure}[!htbp]
    \centering
    \includegraphics[width=0.8\textwidth]{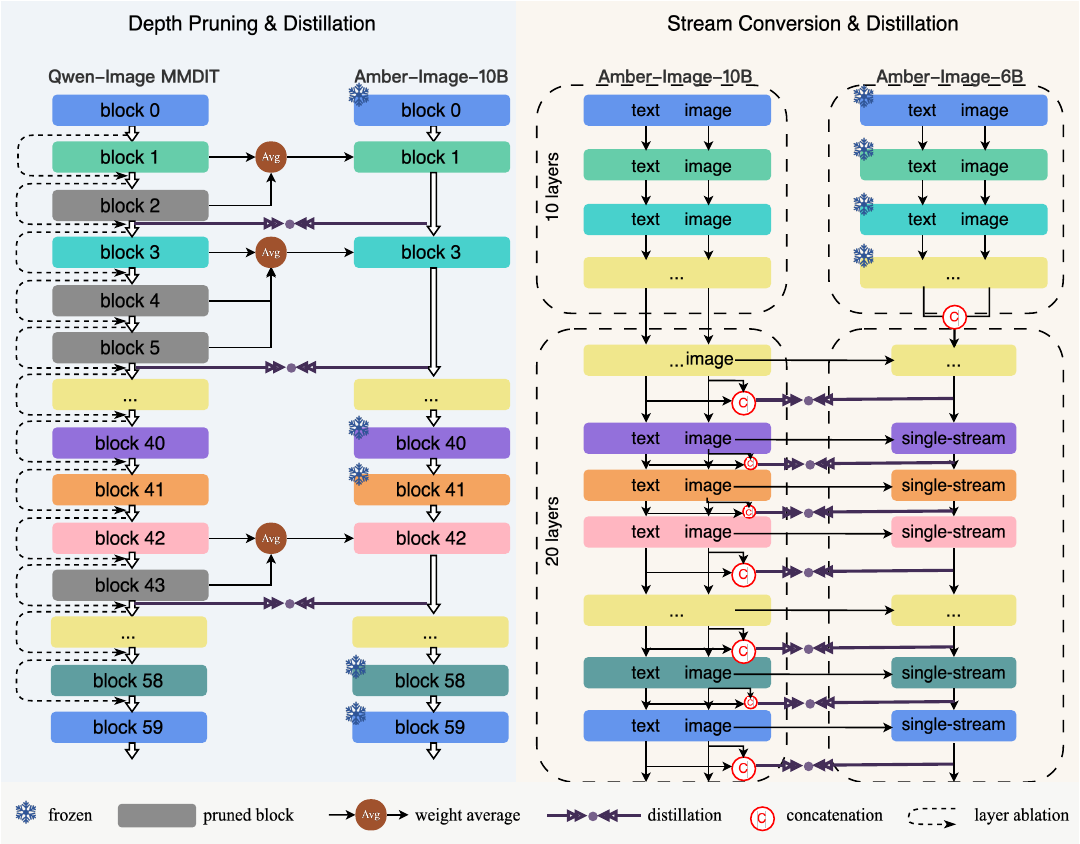}
    \vspace{-0.5em}
    \caption{Overview of the Amber-Image compression pipeline.}
    \vspace{-0.5em}
    \label{fig:Amber-Image_pipeline}
\end{figure}

The overall framework and the hierarchical compression procedure are illustrated in Fig.~\ref{fig:Amber-Image_pipeline}, while the precise execution flows for the two stages are summarized in \textbf{Algorithm~\ref{alg1}}  and \textbf{Algorithm~\ref{alg2}}, respectively. The pipeline consists of two sequential stages: (1) depth pruning to derive a 30-layer model, Amber-Image-10B—where the first and last layers are excluded from pruning due to their essential roles in encoding the raw input and decoding the final output, and (2) deep-layer single-stream conversion to obtain a hybrid-stream variant, Amber-Image-6B, which retains dual-stream processing in early layers and adopts a unified single stream in deeper layers (initialized from the image branch). Each stage employs a two-phase recovery strategy: first, targeted knowledge distillation; second, lightweight full-parameter fine-tuning.

\begin{algorithm}[!htbp]
\small
\SetAlgoLined
\NoCaptionOfAlgo
\caption{\textbf{Algorithm 1:} Depth Pruning and Recovery }
\label{alg1}
\KwIn{Pre-trained MMDiT model $\Theta$ (60 layers), representative prompt set $\mathcal{P}$, subset of timesteps $\mathcal{T}_{sub}$, target depth $K=30$.}
\KwOut{Compressed Amber-Image-10B model.}
\tcp{Stage 1: Layer Importance Estimation}
\For{each layer $l \in \{1, \dots, 58\}$ (excluding input layer 0 and output layer 59)}{
    Initialize cumulative discrepancy $D_l = 0$; \\
    \For{each prompt $p \in \mathcal{P}$}{
        Sample $t \in \mathcal{T}_{sub}$ and latent $z_t$; \\
        Compute original noise $\epsilon_\theta(z_t, t, p)$; \\
        Perform ablation (set $m_l=1$ in Eq. 1) to get $\epsilon_{\theta \setminus \{l\}}(z_t, t, p)$; \\
        Calculate weighted discrepancy $\delta_l = \omega_t \cdot \|\epsilon_\theta - \epsilon_{\theta \setminus \{l\}}\|_2^2$; \\
        $D_l \gets D_l + \delta_l$;
    }
    Calculate importance score $I_l = D_l / (|\mathcal{P}| \cdot \sum \omega_t)$;
}
Identify $\mathcal{L}_{rem}$ as the set of 30 layers with lowest $I_l$ scores; \\
Identify $\mathcal{L}_{keep}$ as the remaining 30 layers;

\tcp{Stage 2: Local Weight Averaging}
\For{each layer $l \in \mathcal{L}_{keep}$}{
    $k \gets$ number of consecutive layers in $\mathcal{L}_{rem}$ immediately following $l$; \\
    $\mathbf{W}_l^{init} \gets \frac{1}{k+1} \sum_{j=0}^{k} \mathbf{W}_{l+j}^{orig}$;
}
\tcp{Stage 3: Recovery}
1. \textbf{Targeted Distillation:} Freeze layers in $\mathcal{L}_{frozen}$ (where $k=0$); Train layers in $\mathcal{L}_{train}$ (where $k>0$) using teacher hidden states; \\
2. \textbf{Global Fine-tuning:} Unfreeze all layers and optimize with standard diffusion loss.
\normalsize
\end{algorithm}

\subsection{Depth Pruning and Recovery for Amber-Image-10B}
\subsubsection{Layer Importance Estimation}
\label{sec:lwa}
To guide structured depth reduction, we estimate the importance of each of the 60 MMDiT layers via a global ablation-based scoring protocol. Unlike gradient-based Taylor approximations \parencite{molchanov2019taylor}, which rely on local linear assumptions and often fail to capture nonlinear inter-layer dependencies in deep transformers, our approach directly measures the impact of layer removal on overall generation fidelity.

Specifically, we define the ablation of the $l$-th layer as the selective truncation of its residual contribution, and measure the resulting change in fidelity (e.g., loss) as a global sensitivity indicator. Formally, for a given input $h_l$, the modified output $\tilde{h}_{l+1}$ is expressed as:
\begin{equation}
    \tilde{h}_{l+1} = (1 - m_l) \cdot \mathcal{F}_l(h_l) + h_l
\end{equation}
where $\mathcal{F}_l$ denotes the $l$-th Transformer block and $m_l \in \{0, 1\}$ is a binary mask. During the assessment of layer $l$, we set $m_l = 1$ (disabling the block) while maintaining $m_{i \neq l} = 0$.

We quantify this effect via the layer-wise prediction discrepancy $\delta_l$, defined in the noise prediction space as the difference between the original and ablated noise predictions:
\begin{equation}
    \delta_l(z_t, t, p) = \epsilon_\theta(z_t, t, p) - \epsilon_{\theta \setminus \{l\}}(z_t, t, p) 
\end{equation}
where $z_t$ denotes the latent representation at timestep $t$, and $\epsilon_\theta, \epsilon_{\theta \setminus \{l\}}$ represent the noise predictions of the original and the model with the $l$-th layer ablated, respectively.

Following observations in Diff-Pruning \parencite{fang2023diffprune} that prediction errors at different timesteps contribute unequally to final generation quality, we employ a dynamic weighting kernel $\omega_t$. We explicitly account for the chained effect inherent in the reverse diffusion process: errors introduced at early timesteps (high-noise) tend to disrupt global semantic structures, whereas those at later timesteps primarily affect fine-grained texture details. Accordingly, we assign higher weights to larger $t$ values to prioritize the preservation of fundamental semantic layouts during assessment. The final importance score $I_l$ is defined as:
\begin{equation} 
I_l(\mathcal{P}) = \frac{1}{|\mathcal{P}|} \sum_{p \in \mathcal{P}} \left( \frac{\sum_{t \in \mathcal{T}_{sub}} \omega_t \cdot \| \delta_l(z_t, t, p) \|_2^2 }{\sum_{t \in \mathcal{T}_{sub}} \omega_t} \right) 
\end{equation}
where $\mathcal{P}$ is a representative prompt set, and $\omega_t$ is a timestep-dependent weighting factor.

The resulting score $I_l(\mathcal{P})$ constitutes our layer importance estimation, which holistically quantifies each layer’s contribution to semantic integrity and visual fidelity. Layers with the lowest $I_l$ scores are identified as redundant and slated for pruning.

\subsubsection{Depth Pruning and Local Weight Averaging}
\label{sec:2wa}
Guided by the importance scores, we compress the 60-layer MMDiT backbone by removing the 30 least important layers, as ranked by their scores $I_l$ from  Section~\ref{sec:lwa}. This stage effectively reduces the model depth by 50$\%$, significantly decreasing the computational overhead.

The local weight average ensures that the preserved layers inherit the transformation capabilities of their pruned neighbors. For any preserved layer $l \in \mathcal{L}_{keep}$, let $k$ denote the number of consecutive layers immediately following $l$ that have been assigned to $\mathcal{L}_{rem}$. We initialize the weights of this preserved layer, $\mathbf{W}_l^{init}$, as the arithmetic mean of the original weights from the cluster of layers it now represents:
\begin{equation}
    \mathbf{W}_l^{init} = \frac{1}{k+1} \sum_{j=0}^{k} \mathbf{W}_{l+j}^{orig}
\end{equation}

For layers where no subsequent pruning occurs (i.e., $k=0$), the weights remain unchanged from their original state ($\mathbf{W}_l^{init} = \mathbf{W}_l^{orig}$). 

The core idea is that adjacent layers in MMDiT typically perform incremental refinements on latent features, and their collective transformation can be well approximated by the local mean in weight space. With this strategy, we effectively mitigate the sharp loss spike commonly observed at the start of training, and provide a high-quality warm start. This significantly accelerates the recovery of generation fidelity compared to random or naive initialization strategies.

\subsubsection{Targeted Distillation and Global Fine-tuning}
To bridge the fidelity gap introduced by layer reduction and weight fusion, we propose a two-stage recovery protocol. This process initiates with selective layer-wise distillation for rapid alignment and concludes with global fine-tuning to further improve generation quality.

Instead of immediate full-parameter optimization, we adopt a distillation strategy that operates only on the newly initialized layers. In the pruned student model, all retained layers from the original architecture are partitioned into two sets: (1) layers reinitialized via local weight averaging (Section~\ref{sec:2wa}), denoted as $l \in \mathcal{L}_{train}$, which are set to trainable; and (2) the remaining retained layers, denoted as $l \in \mathcal{L}_{frozen}$, whose weights are frozen.

To further recover lost representational capacity, we perform layer-wise knowledge distillation using the original 60-layer Qwen-Image as the teacher. Only layers reinitialized via weight averaging are trained; all others remain frozen. For each trainable student layer at original position $i$, we use the hidden state of the teacher’s last layer in the corresponding pruned cluster (i.e., the maximal-index layer among the pruned layers adjacent to the preserved layer $i$) as the distillation target. This encourages the student layer to implicitly reconstruct the cumulative transformation of the entire removed blocks.

Following distillation, we release all parameter constraints and perform a limited number of iterations of full-parameter fine-tuning using the standard diffusion objective. This step harmonizes the heterogeneous components—original retained blocks and reinitialized blocks—into a unified representational space, yielding the final Amber-Image-10B model with superior visual fidelity and semantic consistency.

\subsection{Deep-Layer Single-Stream Conversion and Refinement for Amber-Image-6B}
To further reduce model parameters and improve training and inference efficiency, we convert the deep layers of the dual-stream backbone into a single-stream architecture. Specifically, the first 10 layers remain dual-stream for modality-specific processing, while layers 11--30 are replaced by shared single-stream blocks initialized from the image branch of the 10B teacher. This design addresses architectural redundancy in the MMDiT backbone while retaining the image stream's spatial and generative priors as the starting point for the unified representation.

\begin{algorithm}[H]
\SetAlgoLined
\NoCaptionOfAlgo
\caption{\textbf{Algorithm 2:} Deep-Layer Single-Stream Conversion and Refinement}
\label{alg2}
\KwIn{Amber-Image-10B teacher model $\Theta_{10B}$.}
\KwOut{Amber-Image-6B hybrid-stream model.}

\tcp{Stage 1: Architecture Hybridization}
Maintain layers $1 \dots 10$ as dual-stream blocks; \\
Convert layers $11 \dots 30$ to single-stream blocks; \\

\tcp{Stage 2: Single-Stream Weight Initialization (Image Branch)}
\For{each single-stream layer $l \in \{11, \dots, 30\}$}{
    $\mathbf{W}_{shared} \gets \mathbf{W}_{image}$; \tcp*{where $\mathbf{W} \in \{Q, K, V, O\}$}
}

\tcp{Stage 3: Progressive Alignment and Refinement}
\While{Local Distillation Phase}{
    For layer $l \in [11, 30]$, set $\mathcal{H}_{\text{target}}^{(l)} = \mathrm{Concat}\left( \mathcal{H}_{\text{text}}^{(l,\text{ teacher})},\; \mathcal{H}_{\text{image}}^{(l,\text{ teacher})} \right)$; \\
    Minimize $\mathcal{L}_{MSE} = \| \mathcal{H}_{\text{student}}^{(l)} - \mathcal{H}_{\text{target}}^{(l)} \|_2^2$ with early layers frozen;
}
\textbf{Refinement:} Perform lightweight full-parameter fine-tuning on Amber-Image-6B.
\end{algorithm}

\subsubsection{Hybrid-Stream Architecture and Initialization}
Based on the observation that cross-modal features (text and image) exhibit high semantic correlation in deeper stages, we propose a transition from a dual-stream to a single-stream architecture in the late stages of the backbone. Specifically, for the 30-layer backbone of Amber-Image-10B, we maintain the first 10 layers in their original dual-stream configuration to preserve early-stage modality-specific feature extraction. The subsequent 20 layers are transformed into single-stream blocks, resulting in a "10+20" hybrid-stream architecture. This design strategically balances the need for independent modal processing in early stages with parameter parsimony in deeper stages, reducing the backbone's parameters by approximately $40\%$ and forming the basis for the Amber-Image-6B variant.

The single stream requires one set of weight matrices $\mathbf{W}_{shared}$ per layer. We initialize these directly from the \emph{image}-stream projection weights of the 10B teacher:
\begin{equation}
\mathbf{W}_{shared} = \mathbf{W}_{image}
\end{equation}
where $\mathbf{W} \in \{\mathbf{W}_q, \mathbf{W}_k, \mathbf{W}_v, \mathbf{W}_{out}\}$. We adopt image-side initialization because the image stream is the primary pathway for spatial structure and pixel-level generation; the text modality is then integrated via the subsequent distillation phase, which aligns the single stream to the concatenated teacher hidden states of both streams.

\subsubsection{Progressive Alignment and Refinement }
To bridge the fidelity gap introduced by the single-stream conversion, we perform a two-stage recovery protocol starting with local knowledge distillation. During this phase, the first 10 dual-stream layers of the student Amber-Image-6B are kept frozen to serve as stable semantic anchors. The 20 single-stream layers are then trained to emulate the behavior of the Amber-Image-10B teacher.

The bridge layer at depth 10, which interfaces the dual-stream and single-stream regions, receives a concatenated representation of the image and text hidden states. For all subsequent single-stream layers $l \in [11, 30]$, the distillation supervision signal is derived from the concatenated text and image hidden states of the Amber-Image-10B teacher:
\begin{equation} 
\mathcal{H}_{\text{target}}^{(l)} = \mathrm{Concat}\left( \mathcal{H}_{\text{text}}^{(l,\text{ teacher})},\; \mathcal{H}_{\text{image}}^{(l,\text{ teacher})} \right)
\end{equation}

The student output $\mathcal{H}_{\text{student}}^{(l)}$ is optimized to minimize the MSE loss against this target, ensuring the single-stream block effectively integrates information previously handled by independent branches.

Finally, we conduct a lightweight round of full-parameter fine-tuning using the standard diffusion loss. By optimizing against the standard diffusion objective, the model reconciles the functional transition between the fixed dual-stream and the newly calibrated single-stream regions, delivering the final Amber-Image-6B model with high generative fidelity and significantly reduced parameters.

\section{Experiment}

\subsection{Dataset}
Our dataset consists of approximately one million image–text pairs, entirely derived from internal sources. It comprises two components: a real-world subset and a synthetic subset. The real-world subset covers diverse visual domains, including portraits, landscapes, food, and pets, providing broad coverage of natural scenes and objects. The synthetic subset is specifically constructed to focus on text-rendering scenarios, supplementing the real-world data with cases that emphasize accurate character generation and layout control. Each image is annotated with multiple descriptive captions, typically ranging from three to eight per image, with variations in length, perspective, and phrasing to support robust cross-modal alignment learning.

For layer-importance estimation in the depth pruning stage, we randomly sample 10,000 image–text pairs that span a wide range of concepts and object categories. This subset ensures balanced representation across semantic domains, enabling reliable estimation of layer contributions without incurring excessive computational overhead. For the subsequent knowledge distillation and full-parameter fine-tuning stages, we utilize the complete dataset of one million image–text pairs to maximize data diversity and model performance recovery.

\subsection{Training Setup}
All experiments are conducted on a single node equipped with 8 NVIDIA A100 GPUs. The entire training pipeline for both Amber-Image-10B and Amber-Image-6B proceeds through the following sequential phases:

\begin{enumerate}
    \item \textbf{Layer importance estimation.} We evaluate per-layer sensitivity on 10{,}000 sampled image--text pairs. This stage is inference-only and incurs negligible computational cost.
    \item \textbf{Targeted layer-wise distillation (10B).} Only the reinitialized layers are trained while all other layers remain frozen, running for approximately 24 hours (192 GPU hours).
    \item \textbf{Full-parameter fine-tuning (10B).} All layers of Amber-Image-10B are unfrozen and optimized on the complete one-million-pair dataset for 5 days (960 GPU hours).
    \item \textbf{Local knowledge distillation (6B).} The 20 single-stream layers of Amber-Image-6B are trained against the Amber-Image-10B teacher, with the first 10 dual-stream layers frozen, running for approximately 18 hours (144 GPU hours).
    \item \textbf{Lightweight full-parameter fine-tuning (6B).} All layers of Amber-Image-6B are unfrozen and refined using the standard diffusion objective for 3 days (576 GPU hours).
\end{enumerate}

\noindent The entire pipeline completes in approximately 10 wall-clock days, consuming fewer than 2{,}000 GPU hours in total (1{,}872 GPU hours). This stands in stark contrast to training comparable-scale models from scratch, which typically demands tens of thousands of GPU hours along with carefully engineered multi-stage data pipelines, further underscoring the practical efficiency of our compression-based paradigm.

\subsection{Evaluation}

We evaluate Amber-Image from two complementary perspectives: text rendering fidelity and general text-to-image generation quality. To ensure a rigorous assessment, we utilize a diverse set of established benchmarks. For general generation capability, we employ DPG-Bench \parencite{hu2024dpg} for compositional reasoning, GenEval \parencite{ghosh2023geneval} for semantic alignment, and OneIG-Bench \parencite{chang2025oneigbench} for multi-faceted instruction following. To specifically scrutinize text rendering, we adopt LongText-Bench \parencite{geng2025longtext} and CVTG-2K \parencite{du2025cvtg2k}, which evaluate character integrity and layout coherence in complex, multi-line scenarios.

\subsection{General Text-to-Image Generation}
\noindent \textbf{DPG-Bench.} This benchmark comprises 1,065 dense and semantically rich prompts designed to evaluate a model's adherence to complex instructions. As presented in Table \ref{tab:dpg}, both Amber-Image variants achieve the highest overall scores among all compared models, surpassing not only the closed-source Seedream 3.0 and GPT Image 1, but also the 20B open-source teacher Qwen-Image and all 7B-class open-source competitors including Z-Image, LongCat-Image, and Ovis-Image. Even our more compact Amber-Image-6B closely trails the 10B variant and remains ahead of every baseline. These results demonstrate that our compression framework not only preserves but can even enhance compositional reasoning abilities in dense textual descriptions.

\textbf{GenEval.} To further probe semantic reasoning and object-centric grounding, we evaluate Amber-Image on the GenEval benchmark. As shown in Table \ref{tab:geneval}, both Amber-Image variants achieve the best overall scores among all compared models, outperforming the teacher Qwen-Image, both closed-source systems (Seedream 3.0 and GPT Image 1), and all 7B-class open-source competitors. In particular, our models exhibit notably strong performance in the ``Position'' and ``Attribute'' dimensions, indicating a robust capability for generating spatially-aware, semantically distinct entities. These results underscore that our depth pruning and distillation pipeline effectively preserves---and in some cases strengthens---fine-grained object-centric reasoning.

\textbf{OneIG.} We utilize OneIG-Bench to assess fine-grained alignment and multi-lingual robustness across English (Table \ref{tab:oneig_en}) and Chinese (Table \ref{tab:oneig_zn}). In contrast to the strong results on DPG and GenEval, Amber-Image reveals clear room for improvement on this benchmark. While both variants maintain competitive ``Text'' rendering scores that closely approach the teacher Qwen-Image, a notable gap exists in the ``Style'' and ``Diversity'' dimensions relative to top-performing models such as Z-Image, Ovis-Image, and the closed-source systems. We attribute this gap primarily to the limited scale and diversity of our fine-tuning data, as well as potential aesthetic priors lost during aggressive compression. Incorporating richer stylistic data and reinforcement learning from human feedback (RLHF) are promising directions to mitigate this shortcoming.
\begin{table}[htbp]
\centering
\caption{Quantitative evaluation results on DPG.}
\label{tab:dpg}
\footnotesize
\setlength{\tabcolsep}{2.8pt}
\begin{tabular}{lcccccc}
\toprule
 {\textbf{Model}} &  \textbf{Global} & \textbf{Entity} &  {\textbf{Attribute}} &  {\textbf{Relation}} &  {\textbf{Other}} & \textbf{Overall} \\
\midrule
Seedream 3.0 \parencite{gao2025seedream30}  & \textbf{94.31} & \textbf{92.65} & 91.36 & 92.78 & 88.24 & 88.27  \\
GPT Image 1 \parencite{openai2025gpt4oimage}  & 88.89 & 88.94 & 89.84 & 92.63 & 90.96 & 85.15 \\
\midrule
Emu3-Gen \parencite{wang2024emu3gen} &  85.21 & 86.68 & 86.84 & 90.22 & 83.15 &  80.60 \\
OmniGen2 \parencite{wu2025omnigen2} &  88.81 & 88.83 & 90.18 & 89.37 & 90.27 & 83.57 \\
Janus-Pro \parencite{chen2025januspro}  &  86.90 & 88.90 & 89.40 & 89.32 & 89.48 & 84.19 \\
SD1.5 \parencite{rombach2022latent} &  74.63 & 74.23 & 75.39 & 73.49 & 67.81 & 63.18 \\
SDXL \parencite{podell2023sdxl} &  83.27 & 82.43 & 80.91 & 86.76 & 80.41 & 74.65 \\
PixArt-$\alpha$ \parencite{chen2023pixartalpha} &  74.97 & 79.32 & 78.60 & 82.57 & 76.96 & 71.11 \\
PixArt-$\sigma$ \parencite{chen2024pixartsigma} &  86.89 & 82.89 & 88.94 & 86.59 & 87.68 & 80.54 \\
HiDream-I1-Full \parencite{cai2025hidreami1} &  76.44 & 90.22 & 89.48 & 93.74 & 91.83 & 85.89 \\
SD3 Medium \parencite{esser2024sd3}  & 87.90 & 91.01 & 88.83 & 80.70 & 88.68 & 84.08 \\
FLUX.1[Dev] \parencite{blackforest2024flux1}  &  74.35 & 90.00 & 88.96 & 90.87 & 88.33 & 83.84 \\
HunyuanImage-3.0 \parencite{cao2026hunyuanimage30} & 92.12 & 92.53 & 89.13 & 92.13 & 91.92 & 86.10 \\
Qwen-Image \parencite{wu2025qwenimage} &  91.32 & 91.56 & 92.02 & 94.31 & \textbf{92.73} & 88.32 \\
Z-Image \parencite{imageteam2025zimage}  & \textbf{93.39} & 91.22 & \textbf{93.16} & 92.22 & 91.52 & 88.14 \\
Z-Image-Turbo \parencite{imageteam2025zimage} & 91.29 & 89.59 & 90.14 & 92.16 & 88.68 & 84.86 \\
LongCat-Image \parencite{meituan2025longcat} &  89.10 & \textbf{92.54} & 92.00 & 93.28 & 87.50 & 86.80 \\
Ovis-Image \parencite{wang2025ovisimage} &  82.37 & 92.38 & 90.42 & \textbf{93.98} & 91.20 & 86.59 \\
PPCL-OPPO-10B \parencite{ma2025oppo}	&  85.0	& 86.8 & 	85.6	&  90.5 & 	87.3& 	 81.7 \\
\midrule
Amber-Image-10B &  83.28 & \textbf{92.54} & 90.16 & \textbf{94.47} & 87.60 & \textbf{89.61} \\
Amber-Image-6B & 79.73 & 90.45 & 91.64 & 93.87 & 89.11 & 88.96 \\
\bottomrule
\end{tabular}
\end{table}

\begin{table}[htbp]
\centering
\caption{Quantitative Evaluation results on GenEval.}
\label{tab:geneval}
\footnotesize
\setlength{\tabcolsep}{2.8pt}
\begin{tabular}{lccccccc}
\toprule
{\textbf{Model}} & \textbf{Single} & \textbf{Two} & {\textbf{Counting}} & {\textbf{Colors}} & {\textbf{Position}} & \textbf{Attribute} & {\textbf{Overall}} \\
\midrule
Seedream 3.0 \parencite{gao2025seedream30}  & 0.990 & 0.960 & \textbf{0.910} & \textbf{0.930} & 0.470 & 0.800 & 0.840 \\
GPT Image 1 \parencite{openai2025gpt4oimage}  & 0.990 & 0.920 & 0.850 & 0.920 & 0.750 & 0.610 & 0.840 \\
\midrule
Emu3-Gen \parencite{wang2024emu3gen} &  0.98 & 0.71 & 0.34 & 0.81 & 0.17 & 0.21 & 0.54 \\
OmniGen2 \parencite{wu2025omnigen2} &  1.000 & 0.950 & 0.640 & 0.880 & 0.550 & 0.760 & 0.800 \\
Janus-Pro \parencite{chen2025januspro}  &  0.990 & 0.890 & 0.590 & 0.900 & 0.790 & 0.660 & 0.800 \\
PixArt-$\alpha$ \parencite{chen2023pixartalpha} &  0.98 & 0.5 & 0.44 & 0.8 & 0.08 & 0.07 & 0.48 \\
HiDream-I1-Full \parencite{cai2025hidreami1} &  1.000 & \textbf{0.980} & 0.790 & 0.910 & 0.600 & 0.720 & 0.830 \\
SD3 Medium \parencite{esser2024sd3}  &  0.98 & 0.74 & 0.63 & 0.67 & 0.34 & 0.36 & 0.62 \\
SD3.5-Large \parencite{esser2024sd3}  & 0.98 & 0.89 & 0.73 & 0.83 & 0.34 & 0.47 & 0.71 \\
FLUX.1[Dev] \parencite{blackforest2024flux1}  &  0.98 & 0.81 & 0.74 & 0.79 & 0.22 & 0.45 & 0.66 \\
HunyuanImage-3.0 \parencite{cao2026hunyuanimage30} & 1.00 & 0.920 & 0.48 & 0.820 & 0.420 & 0.630 & 0.720 \\
Qwen-Image \parencite{wu2025qwenimage} &  0.990 & 0.920 & 0.890 & 0.880 & 0.760 & 0.770 & 0.870 \\
Z-Image \parencite{imageteam2025zimage}  & 1.000 & 0.940 & 0.780 & \textbf{0.930} & 0.620 & 0.770 & 0.840 \\
Z-Image-Turbo \parencite{imageteam2025zimage} & 1.000 & 0.950 & 0.770 & 0.890 & 0.650 & 0.680 & 0.820 \\
LongCat-Image \parencite{meituan2025longcat} &  0.990 & \textbf{0.980} & 0.860 & 0.860 & 0.750 & 0.730 & 0.870 \\
Ovis-Image \parencite{wang2025ovisimage} &  1.000 & 0.970 & 0.760 & 0.860 & 0.670 & 0.800 & 0.840 \\
PPCL-OPPO-10B \parencite{ma2025oppo}& 0.968&	0.885&	0.822&	0.840&	0.521&	0.670	&0.784 \\
\midrule
Amber-Image-10B & 0.963 & 0.849 & \textbf{0.900} & 0.862 & 0.850 & \textbf{0.860} & 0.881 \\
Amber-Image-6B &  0.963 & 0.879 & 0.875 & 0.894 & \textbf{0.880} & 0.810 & \textbf{0.883} \\

\bottomrule
\end{tabular}
\end{table}

\begin{table}[htbp]
\centering
\caption{Quantitative evaluation results on OneIG-EN.}
\label{tab:oneig_en}
\footnotesize
\setlength{\tabcolsep}{2.8pt}
\begin{tabular}{lcccccc}
\toprule
{\textbf{Model}} & \textbf{Alignment} & \textbf{Text} & {\textbf{Reasoning}} & {\textbf{Style}} & {\textbf{Diversity}} & {\textbf{Overall}} \\
\midrule
Seedream 3.0 \parencite{gao2025seedream30}  & 0.818 & 0.865 & 0.275 & 0.413 & 0.277 & 0.530 \\
GPT Image 1 \parencite{openai2025gpt4oimage}  & 0.851 & 0.857 & \textbf{0.345} & \textbf{0.462} & 0.151 & 0.533 \\
\midrule
OmniGen2 \parencite{wu2025omnigen2} &  0.804 & 0.680 & 0.271 & 0.377 & 0.242 & 0.475 \\
Janus-Pro \parencite{chen2025januspro}  &  0.553 & 0.001 & 0.139 & 0.276 & 0.365 & 0.267 \\
HiDream-I1-Full \parencite{cai2025hidreami1} &   0.829 & 0.707 & 0.317 & 0.347 & 0.186 & 0.477 \\
Lumina-Image 2.0 \parencite{qin2025luminaimage20}  &   0.819 & 0.106 & 0.270 & 0.354 & 0.216 & 0.353 \\
SD1.5 \parencite{rombach2022latent}  &  0.565 & 0.010 & 0.207 & 0.383 & \textbf{0.429} & 0.319 \\
SDXL \parencite{podell2023sdxl}  &  0.688 & 0.029 & 0.237 & 0.332 & 0.296 & 0.316 \\
SD3.5-Large \parencite{esser2024sd3}  & 0.809 & 0.629 & 0.294 & 0.353 & 0.225 & 0.462 \\
FLUX.1[Dev] \parencite{blackforest2024flux1}  &  0.786 & 0.523 & 0.253 & 0.368 & 0.238 & 0.434 \\
Qwen-Image \parencite{wu2025qwenimage} &  0.882 & 0.891 & 0.306 & \textbf{0.418} & 0.197 & 0.539 \\
Z-Image \parencite{imageteam2025zimage}  & \textbf{0.881} & 0.987 & 0.280 & 0.387 & 0.194 & \textbf{0.546} \\
Z-Image-Turbo \parencite{imageteam2025zimage} & 0.840 & \textbf{0.994} & 0.298 & 0.368 & 0.139 & 0.528 \\
Ovis-Image \parencite{wang2025ovisimage} &  0.858 & 0.914 & \textbf{0.308} & 0.386 & 0.186 & 0.53 \\
PPCL-OPPO-10B \parencite{ma2025oppo} &0.839 &0.860 &0.249 &0.359 &0.121 &0.485 \\
\midrule
Amber-Image-10B &  0.867 & 0.938 & 0.278 & 0.298 & 0.137 & 0.504  \\
Amber-Image-6B & 0.829 & 0.917 & 0.284 & 0.287 & 0.135 & 0.490  \\
\bottomrule
\end{tabular}
\end{table}

\begin{table}[htbp]
\centering
\caption{Quantitative evaluation results on OneIG-ZH.}
\label{tab:oneig_zn}
\footnotesize
\setlength{\tabcolsep}{2.8pt}
\begin{tabular}{lcccccc}
\toprule
{\textbf{Model}} & \textbf{Alignment} & \textbf{Text} & {\textbf{Reasoning}} & {\textbf{Style}} & {\textbf{Diversity}} & {\textbf{Overall}} \\
\midrule
Seedream 3.0 \parencite{gao2025seedream30}  & 0.793 & 0.928 & 0.281 & 0.397 & 0.243 & 0.528 \\
GPT Image 1 \parencite{openai2025gpt4oimage}  & 0.812 & 0.650 & \textbf{0.300} & \textbf{0.449} & 0.159 & 0.474 \\
\midrule
Janus-Pro \parencite{chen2025januspro}  &  0.324 & 0.148 & 0.104 & 0.264 & \textbf{0.358} & 0.240 \\
HiDream-I1-Full \parencite{cai2025hidreami1} &   0.620 & 0.205 & 0.256 & 0.304 & 0.300 & 0.337 \\
Lumina-Image 2.0 \parencite{qin2025luminaimage20}  &   0.731 & 0.136 & 0.221 & 0.343 & 0.240 & 0.334 \\
Qwen-Image \parencite{wu2025qwenimage} &  \textbf{0.825} & 0.963 & 0.267 & \textbf{0.405} & 0.279 & \textbf{0.548} \\
Z-Image \parencite{imageteam2025zimage}  & 0.793 & \textbf{0.988} & 0.266 & 0.386 & 0.243 & 0.535 \\
Z-Image-Turbo \parencite{imageteam2025zimage} & 0.782 & 0.982 & \textbf{0.276} & 0.361 & 0.134 & 0.507 \\
Ovis-Image \parencite{wang2025ovisimage} &  0.805 & 0.961 & 0.273 & 0.368 & 0.198 & 0.521 \\

PPCL-OPPO-10B \parencite{ma2025oppo} & 0.854 & 0.878 &0.268 &0.365 &0.130 &0.499 \\
\midrule
Amber-Image-10B &  0.798 & 0.975 & 0.221& 0.362 & 0.153 & 0.502  \\
Amber-Image-6B & 0.779 & 0.953 & 0.208 & 0.345 & 0.143 & 0.486  \\
\bottomrule
\end{tabular}
\end{table}

\subsection{Text Rendering}

\textbf{LongText.} We further examine the models performance on rendering extended bilingual sequences using LongText-Bench (Table \ref{tab:longtext}). Amber-Image-10B demonstrates strong bilingual text rendering capabilities, outperforming the closed-source Seedream 3.0 on both English and Chinese splits, and significantly surpassing GPT Image 1 on Chinese text rendering. The more compact Amber-Image-6B, despite its aggressive compression, still comfortably exceeds many larger baselines such as OmniGen2 and FLUX.1[Dev]. However, a moderate gap remains compared to the leading 7B-class open-source text-rendering specialists (Z-Image, Ovis-Image) and the teacher Qwen-Image, particularly for the 6B variant. These results suggest that while our compression pipeline preserves the majority of text rendering fidelity, the long-text generation capability degrades more noticeably under deeper architectural simplification. Addressing this through enhanced text-layout tokens or specialized text-rendering fine-tuning remains a key focus for future iterations.

\textbf{CVTG-2K.} To evaluate performance in complex visual text generation, we conduct quantitative experiments on CVTG-2K, which features varying numbers of text regions (Table \ref{tab:cvtg2k}). Amber-Image-10B maintains reasonable word accuracy that remains stable across increasing region counts, and notably achieves the highest CLIPScore among all compared models, indicating strong semantic alignment between the generated text imagery and the input prompts. The 6B variant shows a moderate decline in word accuracy as the number of text regions increases, reflecting the trade-off introduced by the single-stream conversion in deeper layers. While a gap remains compared to specialized text-rendering models such as Ovis-Image and Z-Image, Amber-Image provides a more balanced overall profile, preserving competitive text legibility while excelling in general image generation quality as demonstrated on DPG-Bench and GenEval.

\begin{table}[htbp]
\centering
\caption{Quantitative evaluation results of English text rendering on CVTG-2K.}
\label{tab:cvtg2k}
\footnotesize
\setlength{\tabcolsep}{3.8pt}
\begin{tabular}{lccccccc}
\toprule
\multirow{2}{*}{\textbf{Model}} & 
\multirow{2}{*}{\textbf{NED$\uparrow$}} & 
\multirow{2}{*}{\textbf{CLIPScore$\uparrow$}} & 
\multicolumn{5}{c}{\textbf{Word Accuracy$\uparrow$}} \\
\cmidrule(lr){4-8}
 & & & 2 regions & 3 regions & 4 regions & 5 regions & average$\uparrow$ \\
\midrule
Seedream 3.0 \parencite{gao2025seedream30}    & 0.8537 & 0.7821 & 0.6282 & 0.5962 & 0.6043 & 0.5610 & 0.5924 \\
GPT Image 1 \parencite{openai2025gpt4oimage}  & 0.9478 & 0.7982 & 0.8779 & 0.8659 & 0.8731 & 0.8218 & 0.8569 \\
\midrule
SD3.5 Large \parencite{esser2024sd3}          & 0.8470 & 0.7797 & 0.7293 & 0.6825 & 0.6574 & 0.5940 & 0.6548 \\
FLUX.1 [dev] \parencite{blackforest2024flux1} & 0.6879 & 0.7401 & 0.6089 & 0.5531 & 0.4661 & 0.4316 & 0.4965 \\
Qwen-Image \parencite{wu2025qwenimage}        & 0.9116 & 0.8017 & 0.8370 & 0.8364 & 0.8313 & 0.8158 & 0.8288 \\
Z-Image \parencite{imageteam2025zimage}       & 0.9367 & 0.7969 & 0.9006 & 0.8722 & 0.8652 & 0.8512 & 0.8671 \\
Z-Image-Turbo \parencite{imageteam2025zimage} & 0.9281 & 0.8048 & 0.8872 & 0.8662 & 0.8628 & 0.8347 & 0.8585 \\
Ovis-Image \parencite{wang2025ovisimage}      & \textbf{0.9695} & \textbf{0.8368} & \textbf{0.9248} & \textbf{0.9239} & \textbf{0.9180} & \textbf{0.9166} & \textbf{0.9200} \\
LongCat-Image \parencite{meituan2025longcat}  & 0.9361 & 0.7859 & 0.9129 & 0.8737 & 0.8557 & 0.8310 & 0.8658 \\
\midrule
Amber-Image-10B                                  & 0.8938 & 0.8116 & 0.8791 & 0.8339 & 0.7959 & 0.6952 & 0.8011 \\
Amber-Image-6B                                   & 0.8523 & 0.8047 & 0.8669 & 0.7994 & 0.7200 & 0.6428 & 0.7573 \\
\bottomrule
\end{tabular}
\end{table}

\begin{table}[htbp]
\centering
\caption{Quantitative evaluation results on LongText-Bench.}
\label{tab:longtext}
\footnotesize
\setlength{\tabcolsep}{2.8pt}
\begin{tabular}{lcc}
\toprule
{\textbf{Model}} & \textbf{LongText-Bench-EN} & \textbf{LongText-Bench-ZH} \\
\midrule
Seedream 3.0 \parencite{gao2025seedream30}  & 0.896 & 0.878 \\
GPT Image 1 \parencite{openai2025gpt4oimage}  & \textbf{0.956} & 0.619  \\
\midrule
OmniGen2 \parencite{wu2025omnigen2} &  0.561 & 0.059 \\
Janus-Pro \parencite{chen2025januspro}  &  0.019 & 0.006 \\
HiDream-I1-Full \parencite{cai2025hidreami1} &   0.543 & 0.024 \\
FLUX.1[Dev] \parencite{blackforest2024flux1}  &  0.607 & 0.005  \\
Qwen-Image \parencite{wu2025qwenimage} &  \textbf{0.943} & 0.946 \\
Z-Image \parencite{imageteam2025zimage}  & 0.935 & 0.936 \\
Z-Image-Turbo \parencite{imageteam2025zimage} & 0.917 & 0.926 \\
Ovis-Image \parencite{wang2025ovisimage} &  0.922 & \textbf{0.964} \\
PPCL-OPPO-10B \parencite{ma2025oppo} & 0.871 & 0.885 \\
\midrule
Amber-Image-10B &  0.911 & 0.915 \\
Amber-Image-6B & 0.870 & 0.876 \\
\bottomrule
\end{tabular}
\end{table}

\section{Conclusion}

In this paper, we present Amber-Image, a series of efficient text-to-image models developed using a novel compression framework. By reducing the 60-layer dual-stream MMDiT of Qwen-Image by $70\%$ in parameters, we have successfully developed the Amber-Image-10B and Amber-Image-6B variants without the need for training from scratch or large-scale data engineering. The entire training pipeline consumes fewer than 2{,}000 GPU hours on 8 NVIDIA A100 GPUs, underscoring the exceptional efficiency of our compression-based paradigm. On general text-to-image benchmarks, Amber-Image achieves state-of-the-art results on DPG-Bench and GenEval, surpassing all compared models---including closed-source systems and the significantly larger teacher model---demonstrating that our compression pipeline preserves and even enhances compositional reasoning and semantic grounding. On text rendering tasks (LongText-Bench and CVTG-2K), Amber-Image-10B outperforms several closed-source baselines and maintains competitive fidelity, though a gap remains compared to the best text-rendering specialists. We also observe that the ``Style'' and ``Diversity'' dimensions on OneIG-Bench remain a weakness, which we attribute to the limited diversity of our fine-tuning data and aesthetic priors lost during compression.

To address these limitations, our next step will incorporate reinforcement learning from human feedback (RLHF) to align generation quality more closely with human preferences, particularly in stylistic diversity and aesthetic refinement. Furthermore, we plan to expand the Amber-Image family into various vertical domains by introducing ultra-lightweight specialized models (approximately 2--3B). These domain-specific models will provide tailored, high-efficiency solutions for specialized industrial applications, further demonstrating the scalability and versatility of our compression approach.

\section*{Authors}

\noindent\textbf{Core Contributors:} Chaojie Yang*, Tian Li*, Yue Zhang*, Jun Gao

\vspace{0.5em}
\noindent\small{* Equal contribution.}

\printbibliography

\end{document}